\documentclass[preprint,review,12pt]{elsarticle}

\usepackage{amsmath,amsthm,amssymb,amsfonts,bbm,mathrsfs,algorithmic,algorithm, bm}
\usepackage{array,multirow,multicol,color,xcolor,setspace,caption,subcaption,url,booktabs}
\usepackage{lineno}
\usepackage{verbatim}

\newtheorem*{proof*}{Proof}
\newtheorem*{remark*}{Remark}

\journal{Signal Processing}

\begin{document}
	
	\begin{frontmatter}
		
		
		
		\title{Efficient Tensor Contraction via Fast Count Sketch}
		\author[label]{Xingyu Cao\corref{cor1}}
		\ead{xingyucao@std.uestc.edu.cn}
		\cortext[cor1]{Corresponding author.}
		\author[label]{Jiani Liu}
		\ead{jianiliu@std.uestc.edu.cn}
		\address[label]{School of Information and Communication Engineering, University of Electronic Science and Technology of China (UESTC), Chengdu, 611731, China.}
		
		
		
		\begin{abstract}
			
			Sketching uses randomized Hash functions for dimensionality reduction and acceleration. The existing sketching methods, such as count sketch (CS), tensor sketch (TS), and higher-order count sketch (HCS), either suffer from low accuracy or slow speed in some tensor based applications. In this paper, the proposed fast count sketch (FCS) applies multiple shorter Hash functions based CS to the vector form of the input tensor, which is more accurate than TS since the spatial information of the input tensor can be preserved more sufficiently. When the input tensor admits CANDECOMP/PARAFAC decomposition (CPD), FCS can accelerate CS and HCS by using fast Fourier transform, which exhibits a computational complexity asymptotically identical to TS for low-order tensors. The effectiveness of FCS is validated by CPD, tensor regression network compression, and Kronecker product compression. Experimental results show its superior performance in terms of approximation accuracy and computational efficiency.
		\end{abstract}
		
		\begin{keyword}
			accelerated tensor contraction \sep random projection \sep dimensionality reduction \sep tensor decomposition \sep deep neural network compression 
			
			
		\end{keyword}
		
	\end{frontmatter}
	
	
	\section{Introduction}
	\label{section-introduction}
	Many real-world data exhibit multi-dimensional form and can be naturally represented by tensors. Tensor decomposition is an essential tool for multi-way data analysis \cite{GuhaniyogiR2017BTR, liu2021low, JiLiu2013TCfE, feng2020robust}. As a generalization of the matrix product, tensor contraction is widely used in tensor factorization based methods \cite{AnandkumarA2014TDfL, anandkumar2014guaranteed, wang2015fast, huang2020provable}. However, tensor contraction is time-consuming and costs a lot of memory, especially when the data size is large. To this end, some randomization methods are developed to accelerate tensor decomposition algorithms \cite{ReynoldsMatthewJ.2016RALS, battaglino2018practical, YuanLonghao2019RTRD, MinsterRachel2020RAfL, rakhshan2020tensorized}. 
	
	Sketching adopts this randomization strategy, which succinctly maps the input data into a low-dimensional sketched space while preserving certain data properties. Unlike general random sampling, the sketching techniques commonly use random sparse matrices with certain structure, which are more efficient for computation with guaranteed performance \cite{sarlos2006improved, ClarksonKenneth2017LAaR}. 
	Sketching methods are successfully applied in low-rank approximation \cite{ClarksonKenneth2017LAaR, clarkson2009numerical, WoodruffDavidP.2014SaaT}, regression \cite{sarlos2006improved, ClarksonKenneth2017LAaR, chowdhury2018iterative}, etc.
	
	Charikar et al. \cite{CharikarM2002Ffii} propose a simple and effective sketching method termed Count sketch (CS) to estimate the frequency of items in a data stream. They use a random Hash function pair to map the input vector into a low-dimensional sketched space. Pagh \cite{pagh2013compressed} combines CS of the outer product of two vectors with fast Fourier transform (FFT) to accelerate matrix multiplication compression. Despite the effectiveness of CS, it only applies to vector-valued data. When the input data is a high-order tensor, it needs to vectorize the tensor and generate a long pair of Hash functions to match the dimension of the vectorized tensor. Therefore, the storage cost for the Hash functions is high. 
	
	Pham et al. \cite{PhamNinh2013Fasp} propose tensor sketch (TS) by extending CS to high-dimensional space for polynomial kernel approximation. TS is widely used in multi-dimensional data processing tasks such as tensor decomposition \cite{wang2015fast, yang2018parasketch, malik2018low, sun2020low}, Kronecker product regression \cite{diao2018sketching, diao2019optimal}, and neural network compression \cite{kasiviswanathan2018network, han2020polynomial}. 
	
	Although TS is suitable for tensor data, it inadequately exploits the multi-dimensional structure information of the input tensor by directly sketching it into a vector. To fully exploit the multi-dimensional structure information within tensors, Shi et al. \cite{shi2019efficient} propose higher-order count sketch (HCS), which sketches the input tensor into a lower-dimensional one of the same order. However, HCS suffers from low accuracy and slow speed in some tensor based applications. Therefore, it is essential to develop a new sketching method that can achieve good approximation quality with competitive running speed.
	
	In this paper, we propose a fast count sketch (FCS) which combines the advantages of these three sketching methods. For a general tensor $\mathcal{T}$, FCS applies multiple short Hash functions based CS on ${\rm vec}(\mathcal{T})$ instead of generating a long Hash function pair directly as done by ${\rm CS}({\rm vec}(\mathcal{T}))$. In this way, FCS costs less storage for Hash functions, since only the short ones are stored. The computational complexity of FCS for general tensors is $\operatorname{O}({\rm nnz}(\mathcal{T}))$, where ${\rm nnz(\cdot)}$ represents the number of non-zero elements. Meanwhile, compared with TS, the integrated Hash functions allow FCS to exploit the spatial information within tensors more sufficiently, which results in more accurate estimation, especially when the Hash length of Hash functions is short. Specially, suppose a tensor $\mathcal{T}$ admits CANDECOMP/PARAFAC decomposition (CPD) \cite{CarrollJ.1970Aoid, harshman1970foundations}, FCS can be accelerated by FFT, which exhibits a computational complexity asymptotically identical to TS for low-order tensors, and is much faster than HCS.
	
	To verify the effectiveness of FCS, we apply it to two CPD algorithms named robust tensor power method (RTPM) \cite{AnandkumarA2014TDfL} and alternating least squares (ALS) \cite{KoldaTamaraG.2009TDaA}, which involve two specific tensor contractions $\mathcal{T}(\mathbf{u},\mathbf{u},\mathbf{u})$ and $\mathcal{T}(\mathbf{I},\mathbf{u},\mathbf{u})$. Theoretical proofs guarantee that the estimation by FCS is more accurate than TS when the same Hash functions are used. Experiments on both synthetic and real-world datasets demonstrate FCS achieves better approximation performance with competitive running speed, compared to various counterparts. We also apply FCS to compress the weight tensor of a tensor regression network (TRN) \cite{kossaifi2020tensor} tailed with a CP layer. Experimental results demonstrate FCS achieves better classification performance than other counterparts under various compression ratios. Finally, we compress the Kronecker product and tensor contraction using FCS. Experimental results show that compared to CS, FCS takes less compressing time and Hash memory; compared to HCS, FCS has faster decompressing speed and lower approximation error when the compression ratio is small.

	\section{Background}
	\label{section-background}
	
	\subsection{Notations and basic tensor operations}
	\label{notes}
	Scalars, vectors, matrices and tensors are represented by lowercase, bold lowercase, bold capital and calligraphic letters respectively. Symbols ``$*$'', ``$\circ$'', and ``$\circledast$'', denote Hadamard product, vector outer product, and convolution, respectively. An $N$-th order tensor is rank-$1$ if it can be represented by the outer product of $N$ vectors. Given a tensor $\mathcal{T}\in\mathbb{R}^{I_1\times\cdots\times I_N}$, its vectorization is denoted by ${\rm vec}(\mathcal{T})\in\mathbb{R}^{\prod_{n=1}^{N}I_n}$, its mode-$n$ matricization of is denoted by $\mathbf{T}_{(n)}\in\mathbb{R}^{I_n\times \prod_{i\ne n}I_i}$. The Frobenius norm of $\mathcal{T}$ is represented by $\Vert\mathcal{T}\Vert_{\operatorname{F}}=\Vert{\rm vec}(\mathcal{T})\Vert_{\operatorname{F}}$. For any $N\in\mathbb{N}^+$, we denote $[N]:=\left\{1, \cdots, N\right\}$. For any two tensors $\mathcal{M}, \mathcal{N}$ with the same size, their tensor inner product is denoted by $\left \langle \mathcal{M},\mathcal{N}\right \rangle={\rm vec}(\mathcal{M})^{\operatorname{T}}{\rm vec}(\mathcal{N})$. The CPD of $\mathcal{T}\in\mathbb{R}^{I_1\times\cdots\times I_N}$ is a sum of rank-$1$ tensors, i.e. $\mathcal{T}\approx\sum_{r=1}^R\lambda_r\mathbf{u}^{(1)}_r\circ\cdots\circ\mathbf{u}^{(N)}_r:=[\![\bm{\lambda}; \mathbf{U}^{(1)},\cdots,\mathbf{U}^{(N)}]\!]$, where $R$ is the CP rank, $\bm{\lambda}\in\mathbb{R}^R$, $\mathbf{U}^{(n)}=[\mathbf{u}^{(n)}_1,\cdots,\mathbf{u}^{(n)}_R]\in\mathbb{R}^{I_n\times R}$ for $n\in[N]$. 
	
	Given any two tensors $\mathcal{X}\in\mathbb{R}^{I_1\times I_2\times\cdots\times I_P}$, $\mathcal{Y}\in\mathbb{R}^{J_1\times J_2\times\cdots J_Q}$, their contraction is denoted and computed by $[\mathcal{X}\circledcirc\mathcal{Y}]_{\mathbb{L}}=\mathcal{X}_{\mathbb{P}}\mathcal{Y}_{\mathbb{Q}}=\sum_{\mathbb{O}}\mathcal{X}_{:,\mathbb{O}}\otimes\mathcal{Y}_{\mathbb{O},:}$, where $\mathbb{P}:=\left\lbrace i_1\times\cdots\times i_P\right\rbrace$ for $i_n\in[I_n]$, $\mathbb{Q}:=\left\lbrace j_1\times\cdots\times j_Q\right\rbrace$ for $j_n\in[J_n]$, $\mathbb{O}=\mathbb{P}\cap\mathbb{Q}$ are the specified contraction indices, $\mathbb{L}=(\mathbb{P}\cup\mathbb{Q})\backslash \ \mathbb{O}$ are the free indices, symbol ``$\otimes$'' denotes tensor product. When applied to two vectors or matrices, ``$\otimes$'' degrades to Kronecker product of them. Moreover, given two vectors $\mathbf{u}$ and $\mathbf{v}$, we have ${\rm vec}(\mathbf{u}\circ\mathbf{v})=\mathbf{v}\otimes\mathbf{u}$. Given a series of matrices $\mathbf{M}_n\in\mathbb{R}^{I_n\times J_n}$ for $n\in[N]$, their contraction with a tensor $\mathcal{T}\in\mathbb{R}^{I_1\times\cdots\times I_N}$ is denoted and computed by  $[\mathcal{T}(\mathbf{M}_1,\cdots,\mathbf{M}_N)]_{j_1,\cdots,j_N}=\sum_{i_1=1}^{I_1}\cdots\sum_{i_N=1}^{I_N}\mathcal{T}_{i_1,\cdots,i_N}\mathbf{M}_1(i_1,j_1)\cdots\mathbf{M}_N(i_N,j_N)$ for $j_n\in[J_n]$. Specially, for a $3$rd-order tensor $\mathcal{T}\in\mathbb{R}^{I\times I\times I}$ and a vector $\mathbf{u}\in\mathbb{R}^I$,  $\mathcal{T}(\mathbf{u},\mathbf{u},\mathbf{u})=\left \langle \mathcal{T},\mathbf{u}\circ\mathbf{u}\circ\mathbf{u}\right \rangle$, $\mathcal{T}(\mathbf{I},\mathbf{u},\mathbf{u})_i=\left \langle \mathcal{T},\mathbf{e}_i\circ\mathbf{u}\circ\mathbf{u}\right \rangle$ where $\mathbf{I}\in\mathbb{R}^{I\times I}$ is the identity matrix, $\mathbf{e}_i\in\mathbb{R}^I$ is the $i$th standard basis vector.

	\subsection{Related works}
	In this section, we briefly introduce some sketching techniques that are closely related to our work. 
	
	\emph{Definition 1 (Count Sketch \cite{CharikarM2002Ffii}):} Let $\mathbf{h}:[I]\mapsto [J]$ and $\mathbf{s}:[I]\mapsto\left\{\pm1\right\}$ be two $2$-wise independent Hash functions. The CS of a vector $\mathbf{x}\in\mathbb{R}^I$ takes $\operatorname{O}({\rm nnz}(\mathbf{x}))$ time \cite{ClarksonKenneth2017LAaR} and produces a projection ${\rm CS}(\mathbf{x})\in\mathbb{R}^J$ with each element computed by:
	\begin{equation}
		{\rm CS}(\mathbf{x;\mathbf{h}, \mathbf{s}})_j=\sum_{\mathbf{h}(i)=j}\mathbf{s}(i)\mathbf{x}(i).
		\label{eq:CS}
	\end{equation}When the input is a matrix $\mathbf{X}\in\mathbb{R}^{I\times K}$, CS can be applied on each column of $\mathbf{X}$ and produces a matrix $\mathbf{Y}\in\mathbb{R}^{J\times K}$. Essentially, CS is a sketch for vectors.
	
	\emph{Definition 2 (Tensor Sketch \cite{pagh2013compressed}):} Given an $N$th-order tensor $\mathcal{T}\in\mathbb{R}^{I_1\times\cdots\times I_N}$, $N$ pairs of $2$-wise independent Hash functions $\mathbf{h}_n:[I_n]\mapsto [J], \mathbf{s}_n:[I_n]\mapsto \left\{ \pm1 \right\}$, ${\rm TS}(\mathcal{T})\in\mathbb{R}^J$ is computed in $\operatorname{O}({\rm nnz}(\mathcal{T}))$ time by:	

	\begin{equation}
		{\rm TS}(\mathcal{T};\left\{\mathbf{h}_n, \mathbf{s}_n\right\}_{n=1}^N)_j=\sum_{\mathcal{H}_{i_1,\cdots,i_N}=j}\mathcal{S}_{i_1,\cdots,i_N}\mathcal{T}_{i_1, \cdots, i_N},
		\label{TS_orig}
	\end{equation}where $\mathcal{H}_{i_1,\cdots,i_N}=(\mathbf{h}_1(i_1)+\cdots+\mathbf{h}_N(i_N)-N)\ {\rm mod}\ J + 1$, ${\rm mod}$ denotes the modulo operator, $\mathcal{S}_{i_1,\cdots,i_N}=\mathbf{s}_1(i_1)\mathbf{s}_2(i_2)\cdots\mathbf{s}_N(i_N)$. Specially, when $\mathcal{T}\approx\sum_{r=1}^R\lambda_r\mathbf{u}^{(1)}_r\circ\cdots\circ\mathbf{u}^{(N)}_r:=[\![\bm{\lambda}; \mathbf{U}^{(1)},\cdots,\mathbf{U}^{(N)}]\!]$ is a CP rank-$R$ tensor, ${\rm TS}(\mathcal{T})$ can be computed by the mode-$J$ circular convolution of the CS of each factor vector as: 
	\begin{equation}
		\begin{aligned}
			\label{TS}
			&{\rm TS}(\mathcal{T};\left\{\mathbf{h}_n, \mathbf{s}_n\right\}_{n=1}^N)=\sum_{r=1}^{R}\lambda_r{\rm CS}_1(\mathbf{u}_r^{(1)})\circledast_J\cdots\circledast_J{\rm CS}_N(\mathbf{u}_r^{(N)})\\
			=&\sum_{r=1}^{R}\lambda_r{\operatorname{F}}^{-1}({\operatorname{F}}({\rm CS}_1(\mathbf{U}^{(1)})(:,r))*\cdots*{\operatorname{F}}({\rm CS}_N(\mathbf{U}^{(N)})(:,r))),
		\end{aligned}
	\end{equation}where ${\rm CS}_n(\mathbf{U}^{(n)})\in\mathbb{R}^{J\times R}$ is based on $(\mathbf{h}_n$, $\mathbf{s}_n)$ for $n\in[N]$, $\operatorname{F}$ and ${\operatorname{F}}^{-1}$ denote FFT and its inverse, respectively. By using FFT, (\ref{TS}) can be computed in $\operatorname{O}(\max_n{{\rm nnz}(\mathbf{U}^{(n)})}+RJ{\rm log}J)$ time.
	
	\emph{Definition 3 (Higher-order Count Sketch \cite{shi2019efficient}):} Given $\mathcal{T}\in\mathbb{R}^{I_1\times\cdots\times I_N}$, $N$ pairs of $2$-wise independent Hash functions $\mathbf{h}_n:[I_n]\mapsto [J_n]$, $\mathbf{s}_n:[I_n]\mapsto \left\{ \pm1\right\}$, ${\rm HCS}(\mathcal{T})\in\mathbb{R}^{J_1\times\cdots\times J_N}$ is computed by
	\begin{equation}
		{\rm HCS}(\mathcal{T};\left\{\mathbf{h}_n, \mathbf{s}_n\right\}_{n=1}^N)_{j_1,\cdots,j_N}
		=\sum_{\substack{\mathbf{h}_1(i_1)=j_1\\ \cdots\\ \mathbf{h}_N(i_N)=j_N}} \mathcal{S}_{i_1,\cdots,i_N}\mathcal{T}_{i_1, \cdots, i_N},
		\label{HCS-orig}
	\end{equation}where $\mathcal{S}_{i_1,\cdots,i_N}$ is defined the same as in (\ref{TS_orig}). The computational complexity of (\ref{HCS-orig}) is $\operatorname{O}({\rm nnz}(\mathcal{T}))$. When $\mathcal{T}\approx\sum_{r=1}^{R}\lambda_r\mathbf{u}^{(1)}_r\circ\cdots\circ\mathbf{u}^{(N)}_r=[\![\bm{\lambda}; \mathbf{U}^{(1)}, \cdots, \mathbf{U}^{(N)}]\!]\in\mathbb{R}^{I_1\times\cdots\times I_N}$ is a CP rank-$R$ tensor, ${\rm HCS}(\mathcal{T})$ can be computed by
	\begin{equation}
		{\rm HCS}(\mathcal{T};\left\{\mathbf{h}_n, \mathbf{s}_n\right\}_{n=1}^N)=\sum_{r=1}^{R}\lambda_r{\rm CS}_1(\mathbf{U}^{(1)})(:,r)\circ\cdots\circ{\rm CS}_N(\mathbf{U}^{(N)})(:,r)
		\label{eq:HCS-rank1}.
	\end{equation}Notice that the vector outer product should be materialized when $R>1$. Hence, the computational complexity of (\ref{eq:HCS-rank1}) is $\operatorname{O}(\max_n{{\rm nnz}(\mathbf{U}^{(n)})}$ $+R\prod_{n=1}^{N}J_n)$.
	
	\section{The proposed method}
	As noticed in Section \ref{section-background}, HCS sketches an input tensor into a lower-dimensional one. Therefore, the multi-dimensional structure information can be well preserved. However, for CP rank-$R$ tensors, HCS computes the outer product of CS of each factor vector, which is much slower than TS. Below we will propose FCS, which can achieve a balance between the approximation accuracy and computational complexity.
	
	{\em Definition 4 (Fast Count Sketch):}  Given $\mathcal{T}\in\mathbb{R}^{I_1\times\cdots\times I_N}$, $N$ pairs of $2$-wise independent Hash functions $\mathbf{h}_n:[I_n]\mapsto [J_n]$, $\mathbf{s}_n:[I_n]\mapsto \left\{ \pm1\right\}$, ${\rm FCS}(\mathcal{T})\in\mathbb{R}^{\tilde{J}}$ is defined as:
	\begin{equation}
		{\rm FCS}(\mathcal{T};\left\{\mathbf{h}_n, \mathbf{s}_n\right\}_{n=1}^N):={\rm CS}({\rm vec}(\mathcal{T});\mathbf{h}_{N+1},\mathbf{s}_{N+1}),
		\label{eq:FCS}
	\end{equation}where $\tilde{J}=\sum_{n=1}^{N}J_n-N+1$. $\mathbf{h}_{N+1}:[\tilde{I}]\mapsto [\tilde{J}]$ and $\mathbf{s}_{N+1}: [\tilde{I}]\mapsto \left\{ \pm1 \right\}$ ($\tilde{I}=\prod_{n=1}^{N}I_n$) satisfy:
	\begin{equation}
		\begin{aligned}
			\mathbf{s}_{N+1}(\sum_{n=1}^{N}(i_n-1)\prod_{j=1}^{n-1}I_j+1)&=\prod_{n=1}^{N}\mathbf{s}_n(i_n)\\
			\mathbf{h}_{N+1}(\sum_{n=1}^{N}(i_n-1)\prod_{j=1}^{n-1}I_j+1)&=\sum_{n=1}^{N}\mathbf{h}_n(i_n)-N+1.
		\end{aligned}
	\end{equation}As it does in (\ref{eq:FCS}), FCS on the original tensor equals multiple shorter Hash functions based CS on the vectorized tensor. 
	
	In the next subsections, we will present the computation details of FCS for the CP rank-$R$ tensor and the general tensor. 
	
	\subsection{CP rank-$R$ tensor}
	
	\label{sec:cp rank-R tensor}
	Consider the case where $\mathcal{T}$ admits CPD $\mathcal{T}=\sum_{r=1}^{R}\lambda_r\mathbf{u}^{(1)}_r\circ\cdots\circ\mathbf{u}^{(N)}_r=[\![\bm{\lambda}; \mathbf{U}^{(1)}, \cdots, \mathbf{U}^{(N)}]\!]\in\mathbb{R}^{I_1\times\cdots\times I_N}$. Given $N$ pairs of $2$-wise independent Hash functions $\mathbf{h}_n:[I_n]\mapsto [J_n]$ and $\mathbf{s}_n:[I_n]\mapsto \left\{ \pm1 \right\}$, ${\rm FCS}(\mathcal{T})$ is computed by
	\begin{equation}
		\begin{aligned}
			\label{FCS_syn}
			&{\rm FCS}(\mathcal{T};\left\{\mathbf{h}_n, \mathbf{s}_n\right\}_{n=1}^{N})={\rm CS}(\sum_{r=1}^{R}\lambda_r\overset{1}{\underset{n=N}{\otimes}}(\mathbf{u}^{(n)}_r); \mathbf{h}_{N+1},\mathbf{s}_{N+1})\\
			=&\sum_{r=1}^{R}\lambda_r{\rm CS}_1(\mathbf{U}^{(1)})(:, r)\circledast \cdots\circledast {\rm CS}_N(\mathbf{U}^{(N)})(:, r)\\
			=&\sum_{r=1}^{R}\lambda_r{\operatorname{F}}^{-1}({\operatorname{F}}({\rm CS_1}(\mathbf{U}^{(1)})(:,r), \tilde{J})*\cdots*{\operatorname{F}}({\rm CS}_N(\mathbf{U}^{(N)})(:,r), \tilde{J})),
		\end{aligned}
	\end{equation}where $\overset{1}{\underset{n=N}{\otimes}}(\mathbf{u}^{(n)}_r):=\mathbf{u}^{(N)}_r\otimes\mathbf{u}^{(N-1)}_r\otimes\cdots\otimes\mathbf{u}^{(1)}_r$, $\operatorname{F}(;, \tilde{J})$ and ${\operatorname{F}}^{-1}(;, \tilde{J})$ denote zero-padded $\tilde{J}$-point FFT and its inverse. The proof of (\ref{FCS_syn}) is as follows:
	\begin{proof*}[proof of (\ref{FCS_syn})]
		Given a rank-$1$ tensor $\mathcal{T}=\mathbf{u}^{(1)}\circ\mathbf{u}^{(2)}\circ\cdots\circ\mathbf{u}^{(N)}\in\mathbb{R}^{I_1\times I_2\times\cdots\times I_N}$, if we assign $l=\sum_{n=1}^{N}(i_n-1)\prod_{j=1}^{n-1}I_j+1$, then  ${\rm vec}(\mathcal{T})_l=\mathbf{u}^{(1)}_{i_1}\cdots\mathbf{u}^{(N)}_{i_N}$. We have
			\begin{equation}
				\begin{aligned}
					&\sum_{i_1=1}^{I_1}\cdots\sum_{i_N=1}^{I_N}{\rm vec}(\mathcal{T})_l\mathbf{s}_1(i_1)\cdots\mathbf{s}_N(i_N){\rm w}^{\mathbf{h}_1(i_1)+\cdots+\mathbf{h}_N(i_N)-N}\\
					=&\sum_{i_1=1}^{I_1}\cdots\sum_{i_N=1}^{I_N}\mathbf{u}^{(1)}_{i_1}\cdots\mathbf{u}^{(N)}_{i_N}\mathbf{s}_1(i_1)\cdots\mathbf{s}_N(i_N){\rm w}^{\mathbf{h}_1(i_1)+\cdots+\mathbf{h}_N(i_N)-N}\\
					=&\sum_{i_1}\mathbf{u}^{(1)}_{i_1}\mathbf{s}_1(i_1){\rm w}^{\mathbf{h}_1(i_1)-1}\cdots\sum_{i_N}\mathbf{u}^{(N)}_{i_N}\mathbf{s}_N(i_N){\rm w}^{\mathbf{h}_N(i_N)-1}\\
					=&\mathcal{P}_{{\rm CS}_1(\mathbf{u}^{(1)})}({\rm w})\cdots\mathcal{P}_{{\rm CS}_N(\mathbf{u}^{(N)})}({\rm w}),
				\end{aligned}
			\end{equation}where ${\rm w}$ is a symbolic variable, $\mathcal{P}_{{\rm CS}_n(\mathbf{u}^{(n)})}({\rm w})$ is the polynomial form of ${\rm CS}_n(\mathbf{u}^{(n)})$ for $n\in[N]$. Let $\mathbf{s}_{N+1}(l)=\mathbf{s}_1(i_1)\cdots\mathbf{s}_N(i_N)$, $\mathbf{h}_{N+1}(l)=\mathbf{h}_1(i_1)+\cdots+\mathbf{h}_{N}(i_N)-N+1$, we have
			\begin{equation}
				\begin{aligned}
					&\sum_{i_1=1}^{I_1}\cdots\sum_{i_N=1}^{I_N}{\rm vec}(\mathcal{T})_l\mathbf{s}_1(i_1)\cdots\mathbf{s}_N(i_N){\rm w}^{\mathbf{h}_1(i_1)+\cdots+\mathbf{h}_N(i_N)-N}\\
					=&\sum_l{\rm vec}(\mathcal{T})_l\mathbf{s}_{N+1}(l){\rm w}^{\mathbf{h}_{N+1}(l)-1}\\
					=&\mathcal{P}_{{\rm CS}({\rm vec(\mathcal{T})}; \mathbf{h}_{N+1}, \mathbf{s}_{N+1})}({\rm w}):=\mathcal{P}_{{\rm FCS}(\mathcal{T};\left\{\mathbf{h}_n, \mathbf{s}_n\right\}_{n=1}^N)}({\rm w}).
				\end{aligned}
			\end{equation}Therefore, 
			\begin{equation}
				\mathcal{P}_{{\rm FCS}(\mathcal{T};\left\{\mathbf{h}_n, \mathbf{s}_n\right\}_{n=1}^N)}({\rm w})=\mathcal{P}_{{\rm CS}_1(\mathbf{u}^{(1)})}({\rm w})\cdots\mathcal{P}_{{\rm CS}_N(\mathbf{u}^{(N)})}({\rm w}).
			\end{equation}Due to the fact that polynomial multiplication equals the convolution of their coefficients, we have
			\begin{equation}
				\begin{aligned}
					{\rm FCS}(\mathcal{T};\left\{\mathbf{h}_n, \mathbf{s}_n\right\}_{n=1}^N)&={\rm CS}_1(\mathbf{u}^{(1)})\circledast\cdots\circledast{\rm CS}_N(\mathbf{u}^{(N)})\\
					&=\mathbb{F}^{-1}(\mathbb{F}({\rm CS}_1(\mathbf{u}^{(1)}))*\cdots*\mathbb{F}({\rm CS}_N(\mathbf{u}^{(N)}))).
				\end{aligned}
			\end{equation}Based on the linearity, the conclusion applies to CP rank-$R$ tensors immediately, which completes the proof.
	\end{proof*}
	
	The computational complexity of (\ref{FCS_syn}) is $\operatorname{O}(\max_n{{\rm nnz}(\mathbf{U}^{(n)})}+R\tilde{J}\log \tilde{J})$, which is asymptotically identical to TS for low-order tensors, and is much faster than the vector outer product of HCS shown in (\ref{eq:HCS-rank1}). Since the vectorization form of the CPD is equivalent to the original form after reordering the elements, the multi-dimensional structure of the original tensor can be well preserved.

	\subsection{General tensor}
	
	For most real-world low-rank tensor data, the rank-$1$ factors cannot be known in advance. 
	FCS for general tensors can be computed in $\operatorname{O}({\rm nnz}(\mathcal{T}))$ time by:
		\begin{equation}
			\begin{aligned}
				{\rm FCS}(\mathcal{T}; \left\{\mathbf{h}_n, \mathbf{s}_n\right\}_{n=1}^{N})_j&=\sum_{\mathbf{h}_{N+1}(i)=j}^{\tilde{I}}\mathbf{s}_{N+1}(i)*{\rm vec}(\mathcal{T})_i\\
				&=\sum_{\mathcal{H}_{i_1,\cdots,i_N}=j}\mathcal{S}_{i_1,\cdots,i_N}\mathcal{T}_{i_1, \cdots, i_N},
			\end{aligned}
			\label{FCS_asyn}
		\end{equation}where $\mathcal{H}_{i_1,\cdots,i_N}=\mathbf{h}_1(i_1)+\cdots+\mathbf{h}_N(i_N)- N + 1$,  $\mathcal{S}_{i_1,\cdots,i_N}=\mathbf{s}_1(i_1)\mathbf{s}_2(i_2)\cdots\mathbf{s}_N(i_N)$ for $j\in[\tilde{J}]$. Notice that $\mathcal{H}$ and $\mathcal{S}$ are not necessarily explicitly built so that the only $\left\{\mathbf{h}_n, \mathbf{s}_n\right\}_{n=1}^{N}$ are required to be stored.
	
	We summarize the main differences of FCS among CS, TS, and HCS below:
	
	\begin{enumerate}[(1)]
		\item Compared with CS which applies on ${\rm vec}(\mathcal{T})$, FCS is based on multiple shorter Hash function pairs $\left\{\mathbf{h}_n, \mathbf{s}_n\right\}_{n=1}^N$, while CS generates a long Hash function pair $\mathbf{h}:[\tilde{I}]\mapsto [\tilde{J}]$  and $\mathbf{s}: [\tilde{I}]\mapsto \left\{ \pm1 \right\}$ to match the dimensionality of ${\rm vec}(\mathcal{T})$. Therefore, only $\left\{\mathbf{h}_n, \mathbf{s}_n\right\}_{n=1}^N$ are stored for FCS, which reduces the storage complexity from $\operatorname{O}(\tilde{I})$ to $\operatorname{O}(\sum_{n=1}^{N}I_n)$. Besides, as shown in (\ref{FCS_syn}), FCS for CP rank-$R$ tensors can be computed by FFT, which accelerates the original CS.
		\item Compared with TS, FCS avoids the modular operation in (\ref{TS_orig}) and (\ref{TS}). Intuitively, the multiple Hash functions of FCS are integrated in a way that the spatial structure of the input tensor can be preserved more sufficiently, while the modular operation in TS breaks the spatial relationship of tensors. We will show that FCS provides a more accurate estimator for tensor inner product than TS under the same Hash functions in Proposition $1$. 
		\item Compared with HCS, the vector outer product in (\ref{eq:HCS-rank1}) is also avoided. Besides, as shown in Table \ref{tab:compare sketches}, the approximation for $\mathcal{T}(\mathbf{I},\mathbf{u},\mathbf{u})$ and $\mathcal{T}(\mathbf{u},\mathbf{u},\mathbf{u})$ by FCS is more efficient than HCS under the same Hash length. Finally, we will show in the experiments that FCS has faster decompressing time than HCS for Kronecker product and tensor contraction compression.
	\end{enumerate}
	
	\subsection{Tensor contraction approximation}
	We approximate two specific tensor contractions $\mathcal{T}(\mathbf{u},\mathbf{u},\mathbf{u})$ and $\mathcal{T}(\mathbf{I},\mathbf{u},\mathbf{u})$ using FCS. For brevity, we omit the Hash function pairs in (\ref{TS_orig}) and (\ref{eq:FCS}). The following proposition states the validity of the approximation:
	
	{\em Proposition $1$}: For any two tensors $\mathcal{M}, \mathcal{N}\in\mathbb{R}^{I_1\times I_2\times I_3}$, $\left \langle {\rm FCS}(\mathcal{M}), {\rm FCS}(\mathcal{N})\right \rangle$ provides a consistent estimator of $\left \langle \mathcal{M},\mathcal{N}\right \rangle$ with the variance satisfying
		\begin{equation}
			\label{theorem}
			\operatorname{V}_{\mathbf{h},\mathbf{s}}[\left \langle {\rm FCS}(\mathcal{M}), {\rm FCS}(\mathcal{N})\right \rangle]\le\operatorname{V}_{\mathbf{h},\mathbf{s}}[\left \langle {\rm TS}(\mathcal{M}), {\rm TS}(\mathcal{N})\right \rangle],
		\end{equation}if the Hash functions for TS and FCS are equalized. The validity is guaranteed from two aspects: the consistency guarantees the approximation is feasible, and (\ref{theorem}) illustrates FCS computes a more accurate estimator for tensor inner product than TS \cite{wang2015fast}. The proof is defered to the supplementary materials due to space limit. 
	
	Based on Proposition $1$, the following corollary gives the approximation error:	
	
	{\em Corollary 1}: Given any $3$rd-order tensor $\mathcal{T}\in\mathbb{R}^{I\times I\times I}$, unit vector $\mathbf{u}\in\mathbb{R}^I$, from Chebychev's inequality, if we run the sketch $D=\Omega(\log(1/\delta))$ times, then for any $\epsilon>0$, with probability at least $1-\delta$ we have
		\begin{equation}
			\label{cheby}
			\begin{aligned}
				&\operatorname{P}_{\mathbf{h}, \mathbf{s}}[|\left \langle {\rm FCS}(\mathcal{T}), {\rm FCS}(\mathbf{u}\circ\mathbf{u}\circ\mathbf{u})\right \rangle-\mathcal{T}(\mathbf{u},\mathbf{u},\mathbf{u})|\ge\epsilon]\le \operatorname{O}(\frac{\Vert\mathcal{T}\Vert_\text{F}^2}{J\epsilon^2})\\
				&\operatorname{P}_{\mathbf{h}, \mathbf{s}}[|\left\langle {\rm FCS}(\mathcal{T}), {\rm FCS}(\mathbf{e}_i\circ\mathbf{u}\circ\mathbf{u})\right \rangle-\mathcal{T}(\mathbf{e}_i,\mathbf{u},\mathbf{u})|\ge\epsilon]\le \operatorname{O}(\frac{\Vert\mathcal{T}\Vert_\text{F}^2}{J\epsilon^2}),
			\end{aligned}
		\end{equation}where the symbol $\operatorname{P}$ stands for probability. The Hash lengths of all Hash functions are set to $J$ for simplicity. The proof can be found in the supplementary materials.
	
	\section{Experiments}
	In this section, we verify the effectiveness of the proposed FCS by comparing it with CS, TS and HCS in various numerical experiments, including CPD, TRN compression, the Kronecker product and tensor contraction compression. To make the estimation more robust, we compute $D$ number of independent sketches and return the median for all sketching methods.
	\subsection{CP decomposition}
	We consider two CPD algorithms dubbed RTPM and ALS which involve $\mathcal{T}(\mathbf{u},\mathbf{u},\mathbf{u})$ and $\mathcal{T}(\mathbf{I},\mathbf{u},\mathbf{u})$. From Corollary $1$, the two tensor contractions can be approximated by:
		\begin{equation}
			\label{eq:T_uvw}
			\begin{aligned}
				&\mathcal{T}(\mathbf{u},\mathbf{u},\mathbf{u})\approx\left\langle{\rm FCS}(\mathcal{T}), {\rm FCS}(\mathbf{u}\circ\mathbf{u}\circ\mathbf{u})\right \rangle\\
				=&\left\langle{\rm FCS}(\mathcal{T}), {\operatorname{F}}^{-1}({\operatorname{F}}({\rm CS}_1(\mathbf{u}))*{\operatorname{F}}({\rm CS}_2(\mathbf{u}))*{\operatorname{F}}({\rm CS}_3(\mathbf{u})))\right\rangle
			\end{aligned}
		\end{equation}
		\begin{equation}
			\label{eq:T_Ivw}
			\begin{aligned}
				&\mathcal{T}(\mathbf{I},\mathbf{u},\mathbf{u})_i\approx\left\langle{\rm FCS}(\mathcal{T}), {\rm FCS}(\mathbf{e}_i\circ\mathbf{u}\circ\mathbf{u})\right \rangle\\
				=&\left\langle{\operatorname{F}}^{-1}({\operatorname{F}}({\rm FCS}(\mathcal{T}))*\overline{{\operatorname{F}}({\rm CS}_2(\mathbf u))}*\overline{{\operatorname{F}}({\rm CS}_3(\mathbf u))}),{\rm CS}_1(\mathbf{e}_i)\right \rangle\\
				:=&\left \langle \mathbf{z},{\rm CS}_1(\mathbf{e}_i) \right \rangle
				=\mathbf{s}_1(i)\mathbf{z}(\mathbf{h}_1(i)).
			\end{aligned}
		\end{equation}(\ref{eq:T_Ivw}) holds due to the unitary property of FFT. Both ${\rm FCS}(\mathcal{T})$ and $\mathbf{z}$ are irrelevant to $i$ and can be computed beforehand. We summarize the computational and storage complexity of RTPM and ALS with and without sketching (which we term ``plain'') in Table \ref{tab:compare sketches}. All Hash lengths are set to $J$ for simplicity. 
	\begin{table*}[htbp]
		\centering
		\caption{Computational and storage complexity of the plain, CS, TS, HCS, and FCS based RTPM and ALS for a $3$rd-order tensor $\mathcal{T}\in\mathbb{R}^{I\times I\times I}$. $R$ denotes the target CP rank. }
		\scalebox{0.47}{
			\begin{tabular}{cccccc}
				\toprule
				RTPM/ALS   & plain     & CS     & TS    & HCS		& FCS\\
				\midrule    
				preprocessing and sketch building for $\mathcal{T}\!=\![\![\bm{\lambda};\!\mathbf{U}\!,\!\mathbf{U}\!,\!\mathbf{U}]\!]$ &$\operatorname{O}(RI^3)$             &$\operatorname{O}(RI^3+{\rm nnz}(\mathcal{T}))$           &$\operatorname{O}({\rm nnz}(\mathbf{U})+RJ{\rm log}J)$			&$\operatorname{O}({\rm nnz}(\mathbf{U})+RJ^3)$	&$\operatorname{O}({\rm nnz}(\mathbf{U})+RJ\log J)$		\\
				\specialrule{0pt}{2pt}{2pt}
				preprocessing and sketch building for general tensor $\mathcal{T}$  &-            &$\operatorname{O}({\rm nnz}(\mathcal{T}))$           &$\operatorname{O}({\rm nnz}(\mathcal{T}))$  &$\operatorname{O}({\rm nnz}(\mathcal{T}))$		&$\operatorname{O}({\rm nnz}(\mathcal{T}))$\\
				\specialrule{0pt}{2pt}{2pt}
				$\mathcal{T}(\mathbf{I},\mathbf{u},\mathbf{u})$ or its approximation & $\operatorname{O}(I^3)$          & $\operatorname{O}({\rm nnz}(\mathbf{u})^2I)$            &$\operatorname{O}({\rm nnz}(\mathbf{u})+J\log J + I)$ & $\operatorname{O}({\rm nnz}(\mathbf{u})+IJ^2)$	&$\operatorname{O}({\rm nnz}(\mathbf{u})+J\log J+I)$\\
				\specialrule{0pt}{2pt}{2pt}
				$\mathcal{T}(\mathbf{u},\mathbf{u},\mathbf{u})$ or its approximation   &$\operatorname{O}(I^3)$            &$\operatorname{O}({\rm nnz}(\mathbf{u})^3)$           &$\operatorname{O}({\rm nnz}(\mathbf{u})+J\log J)$	&$\operatorname{O}({\rm nnz}(\mathbf{u})+J^3)$			&$\operatorname{O}({\rm nnz}(\mathbf{u})+J\log J)$\\
				\specialrule{0pt}{2pt}{2pt}
				storage for Hash functions   &-           &$\operatorname{O}(I^3)$           &$\operatorname{O}(I)$ &$\operatorname{O}(I)$	&$\operatorname{O}(I)$\\
				\bottomrule
			\end{tabular}
		}
		\label{tab:compare sketches}
	\end{table*}
	
	\subsubsection{Robust tensor power method}
	RTPM decomposes a noisy input tensor into rank-$1$ components with orthogonal basis vectors. For each rank-$1$ factor, it computes power iteration $\mathbf{u}=\frac{\mathcal{T}(\mathbf{I}, \mathbf{u}, \mathbf{u})}{\Vert \mathcal{T}(\mathbf{I}, \mathbf{u}, \mathbf{u})\Vert_{\operatorname{F}}}$ from random initializations and obtains the eigenvalue by $\mathcal{T}(\mathbf{u}, \mathbf{u}, \mathbf{u})$. 
	
	It is necessary to note that most real-world low-rank tensor data are asymmetric. The power iteration of asymmetric RTPM can be performed similarly as the symmetric case via alternating rank-$1$ update \cite{AnandkumarA2017ATPM}. By (\ref{FCS_asyn}), we can accelerate RTPM on real-world data using FCS.
	
	The effectiveness of FCS is verified by accelerating RTPM on synthetic and real-world datasets. For the synthetic case, we generate a symmetric CP rank-$10$ tensor $\mathcal{T}=\sum_{r=1}^{10}\mathbf{u}_r\circ\mathbf{u}_r\circ\mathbf{u}_r\in\mathbb{R}^{100\times 100 \times 100}$, where $\left\{\mathbf{u}_r\right\}_{r=1}^{10}$ forms a random orthonormal basis. $\mathcal{T}$ is then perturbed by zero-mean Gaussian noise with standard deviation $\sigma=0.01$. The number of independent sketches $D$, initial vectors $L$ and power iterations $T$ are set to $2$, $15$ and $20$, respectively. The Hash functions for TS and FCS are equalized to produce identical initializations. We choose residual norm as the performance metric. 
	
\begin{figure}[H]
		\centerline{\includegraphics[width=\columnwidth]{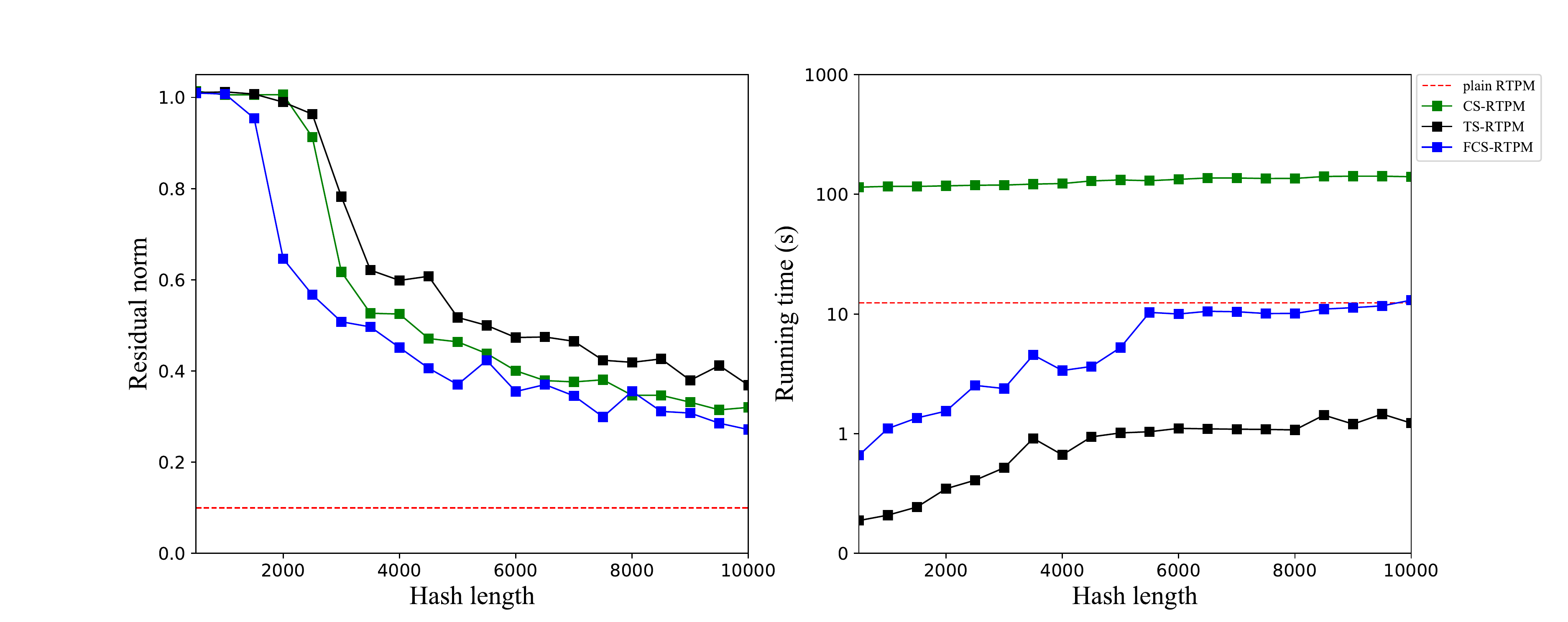}}
		\caption{Performance comparison on a synthetic symmetric CP rank-$10$ tensor $\mathcal{T}\in\mathbb{R}^{100\times100\times100}$ for plain, CS, TS and FCS based RTPM.}
		\label{fig:compare-CS-TS-FCS}
	\end{figure}

	We compare the performance of FCS against the plain, CS, and TS based RTPM with the Hash lengths ranging from $1000$ to $10000$, and the results are shown in Fig. \ref{fig:compare-CS-TS-FCS}. Clearly, FCS-RTPM achieves better approximation accuracy than CS- and TS- RTPM. Although FCS-RTPM is slower than TS-RTPM, it is still much faster than the CS- and plain RTPM under most Hash lengths. Notice that CS-RTPM is even slower than the plain RTPM, hence we do not compare it with FCS-RTPM in the real-world experiment.
	We compare the performance of HCS and FCS based RTPM on a synthetic symmetric CP rank-$10$ tensor $\mathcal{T}\in\mathbb{R}^{50\times50\times50}$. Denote the Hash lengths of HCS and FCS as $J_1$ and $J_2$, respectively. Given a $3$rd-order tensor with dimension $I$, $J_1$ should be smaller than $I$ to sketch the tensor into a lower dimensional one. However, the sketched dimension for FCS requires to be $\operatorname{O}(I^3)$ for provable approximation \cite{wang2015fast}. As a result, it is not fair to set $J_1\approx J_2$ in this experiment. Therefore we choose $J_1$ and $J_2$ such that the sketched dimensions are similar ($J_1^3\approx3J_2-2$). The experiment results are displayed in Table \ref{tab:HCS-FCS-same-memory}. It can be seen that FCS beats HCS in terms of approximation accuracy and running speed under various Hash lengths, noise intensities, and number of independent sketches. 
	
	\newcommand{\tabincell}[2]{\begin{tabular}{@{}#1@{}}#2\end{tabular}}
	\begin{table*}[htbp]
		\begin{center}
			\caption{Performance comparison on a synthetic symmetric CP rank-$10$ tensor $\mathcal{T}\in\mathbb{R}^{50\times50\times50}$ by HCS and FCS based RTPM under similar sketched dimension.}\label{tab:HCS-FCS-same-memory}
			\scalebox{0.58}{
				\begin{tabular}{ccccccccccccc}
					\toprule
					$\sigma$ &              &             & \multicolumn{5}{c}{Residual norm} & \multicolumn{5}{c}{Running time (s)}\\
					\midrule
					\specialrule{0pt}{2pt}{2pt}
					\multirow{11}{*}{0.01} &               & $J_1$ & 14          & 18   & 21   & 23   & 25   & 14          & 18   & 21   & 23   & 25     \\
					\specialrule{0pt}{2pt}{2pt}
					& \multirow{3}{*}{\tabincell{c}{HCS-\\RTPM}}       & $D=10$       & {1.3020}	& {0.8305}	& {0.7744}	& {0.7727}	& {0.7719}	& {12.2329}	& {30.7566}	& {46.7719}	& {58.4108}	& {67.7999}\\
					\specialrule{0pt}{2pt}{2pt}
					&               & $D=15$    &   {0.8237}	& {0.6583}	& {0.6089}	& {0.5938}	& {0.5699}	& {18.2351}	& {45.5026}	& {70.3092}	& {88.6696}	& {104.4252}\\
					\specialrule{0pt}{2pt}{2pt}
					&               & $D=20$      & {0.6904}	& {0.6702}	& {0.5229}	&{0.4738} 	&  {0.4733}	& {24.5617}	& {63.1505}	& {95.1202}	& {119.8723}	& {140.5284}\\
					\specialrule{0pt}{2pt}{2pt}
					&               & $J_2$ & 200          & 250   & 300   & 350   & 400   & 200          & 250   & 300   & 350   & 400    \\
					\specialrule{0pt}{2pt}{2pt}
					& \multirow{3}{*}{\tabincell{c}{FCS-\\RTPM}}     & $D=10$      & \textbf{0.3304}	& \textbf{0.2033}	& \textbf{0.1701}	& \textbf{0.1525}	& \textbf{0.1375}	& \textbf{4.9459}	& \textbf{8.8239}	& \textbf{16.7354}	& \textbf{21.9275}	& \textbf{28.4605}\\
					\specialrule{0pt}{2pt}{2pt}
					&               & $D=15$      & \textbf{0.2440}	& \textbf{0.1794}	& \textbf{0.1472}	& \textbf{0.1280}	& \textbf{0.1179}	& \textbf{6.9054}	& \textbf{13.9064}	& \textbf{26.1508}	& \textbf{34.1260}	& \textbf{42.7100}
					\\
					\specialrule{0pt}{2pt}{2pt}
					&               & $D=20$       &\textbf{0.2135}	& \textbf{0.1544}	& \textbf{0.1226}	& \textbf{0.1050}	& \textbf{0.0899}	& \textbf{8.7913}	& \textbf{18.2077}	& \textbf{34.4879}	& \textbf{44.7941}	& \textbf{56.2395}
					\\
					\specialrule{0pt}{2pt}{2pt}
					\midrule
					\specialrule{0pt}{2pt}{2pt}
					\multirow{11}{*}{0.1}     &          & $J_1$ & 14          & 18   & 21   & 23   & 25   & 14          & 18   & 21   & 23   & 25  \\
					\specialrule{0pt}{2pt}{2pt}
					& \multirow{4}{*}{\tabincell{c}{HCS-\\RTPM}}       & $D=10$       &{0.8941}	& {0.8414}	& {0.7493}	& {0.6796}	& {0.6334}	& 12.1630	& 28.7772	& 42.9283	& 54.9871	& 65.9807
					\\
					\specialrule{0pt}{2pt}{2pt}
					&               & $D=15$       &{0.7692}	& {0.6618}	& {0.6112}	& {0.5766}	& {0.5738}	& 18.0038	& 43.0994	& 65.3426	& 85.6515	& 103.2082
					\\
					\specialrule{0pt}{2pt}{2pt}
					&               & $D=20$       &0.6814	& {0.6155}	& {0.5256}	& {0.4697}	& {0.5409}	& 24.2827	& 57.0479	& 88.6810	& 122.2182	& 149.7220
					\\
					\specialrule{0pt}{2pt}{2pt}
					&               & $J_2$ & 200          & 250   & 300   & 350   & 400   & 200          & 250   & 300   & 350   & 400     \\
					\specialrule{0pt}{2pt}{2pt}
					& \multirow{4}{*}{\tabincell{c}{FCS-\\RTPM}}     & $D=10$       &\textbf{0.3123}	& \textbf{0.2052}	& \textbf{0.1648}	& \textbf{0.1386}	& \textbf{0.1277}	& \textbf{4.8647}	& \textbf{8.7281}	& \textbf{16.5311}	& \textbf{21.5956}	& \textbf{28.6445}
					\\
					\specialrule{0pt}{2pt}{2pt}
					&               & $D=15$       &\textbf{0.2613}	& \textbf{0.1665}	& \textbf{0.1568}	& \textbf{0.1289}	& \textbf{0.1063}	& \textbf{6.8684}	& \textbf{14.6268}	& \textbf{27.2106}	& \textbf{35.0255}	& \textbf{44.4975}
					\\
					\specialrule{0pt}{2pt}{2pt}
					&               & $D=20$       &\textbf{0.2102}	& \textbf{0.1550}	& \textbf{0.1173}	& \textbf{0.0987}	& \textbf{0.0939}	& \textbf{8.9151}	& \textbf{18.8304}	& \textbf{34.3682}	& \textbf{48.4351}	& \textbf{59.3935}
					\\
					\specialrule{0pt}{2pt}{2pt}
					\bottomrule
				\end{tabular}
			}
		\end{center}
	\end{table*}

	For the real-world data based experiment, we compute the plain, TS and FCS based RTPM using the same Hash functions on a hyperspectral imaging (HSI) dataset Watercolors and a light field dataset Buddha. Below we briefly introduce the two datasets and the preprocessing methods:
	\begin{enumerate}[(1)]
		\item The Watercolors is a HSI dataset captured by a resolution $512\times512\times3$ cooled CCD camera with the wevelength ranging from $400$ to $700$ nm at the interval of $10$ ($31$ bands in total) \cite{yasuma2010generalized}. We transform the raw images to gray scale images and represent them as a $512\times512\times31$ tensor. The target CP rank is set as $15$.
		\item The Buddha dataset is captured by a resolution $768\times768\times3$ camera at $9\times9$ views \cite{wanner2013datasets}. Therefore, the raw data can be represented by a $768\times768\times3\times9\times9$ tensor. We transform the raw images to gray scale images, and resize them as a $192\times192\times81$ tensor. The target CP rank is set as $30$.
	\end{enumerate}
	
\begin{figure}[H]
		\centering
		\includegraphics[width=\columnwidth]{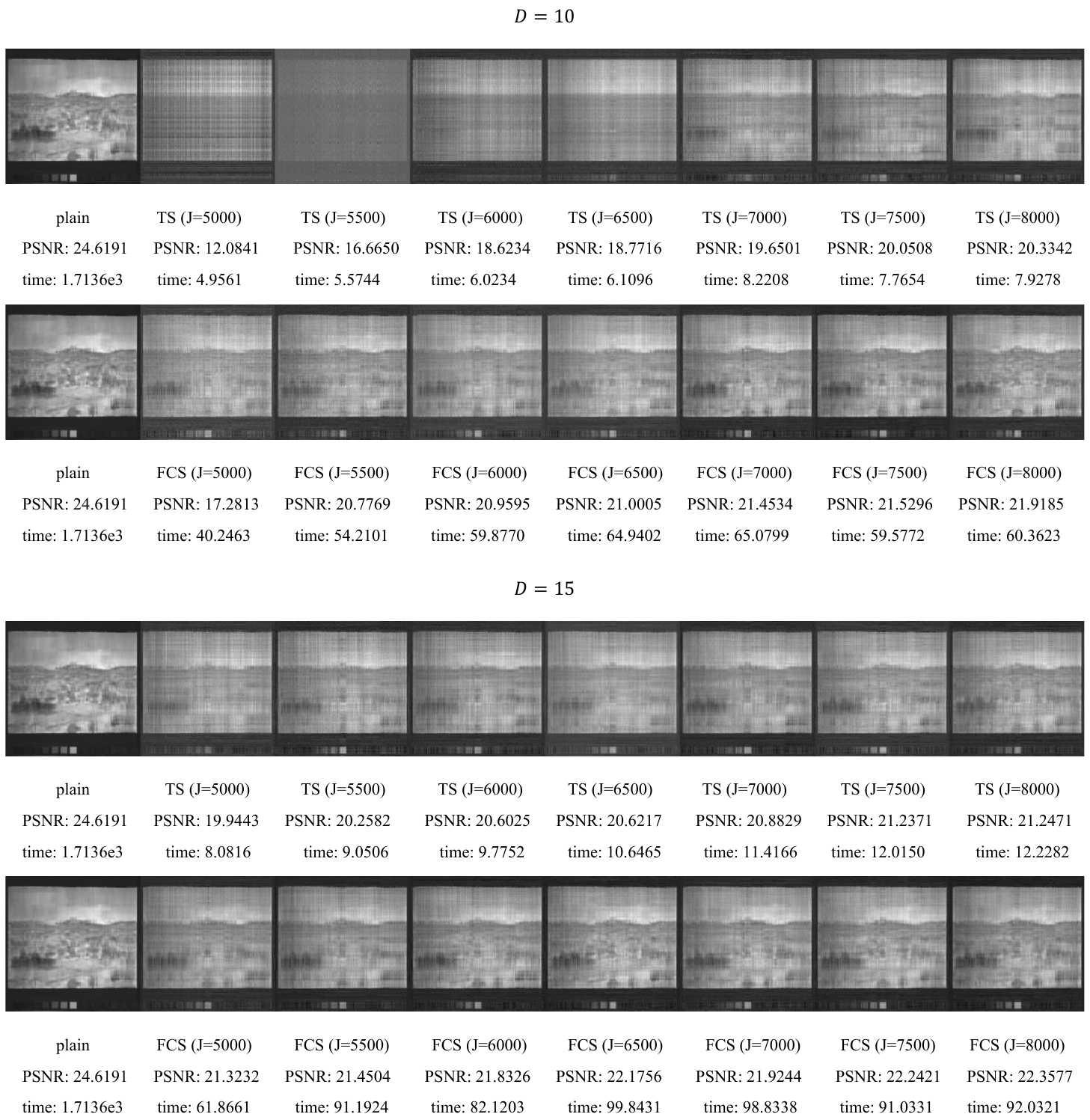}
		\caption{Rank-$15$ approximation of TS-RTPM and FCS-RTPM on HSI dataset Watercolors. PSNR (dB) and running time (s) are tagged under each approximation figure.}
		\label{fig:Watercolors}
	\end{figure}

\begin{figure}[H]
		\centering
		\includegraphics[width=\columnwidth]{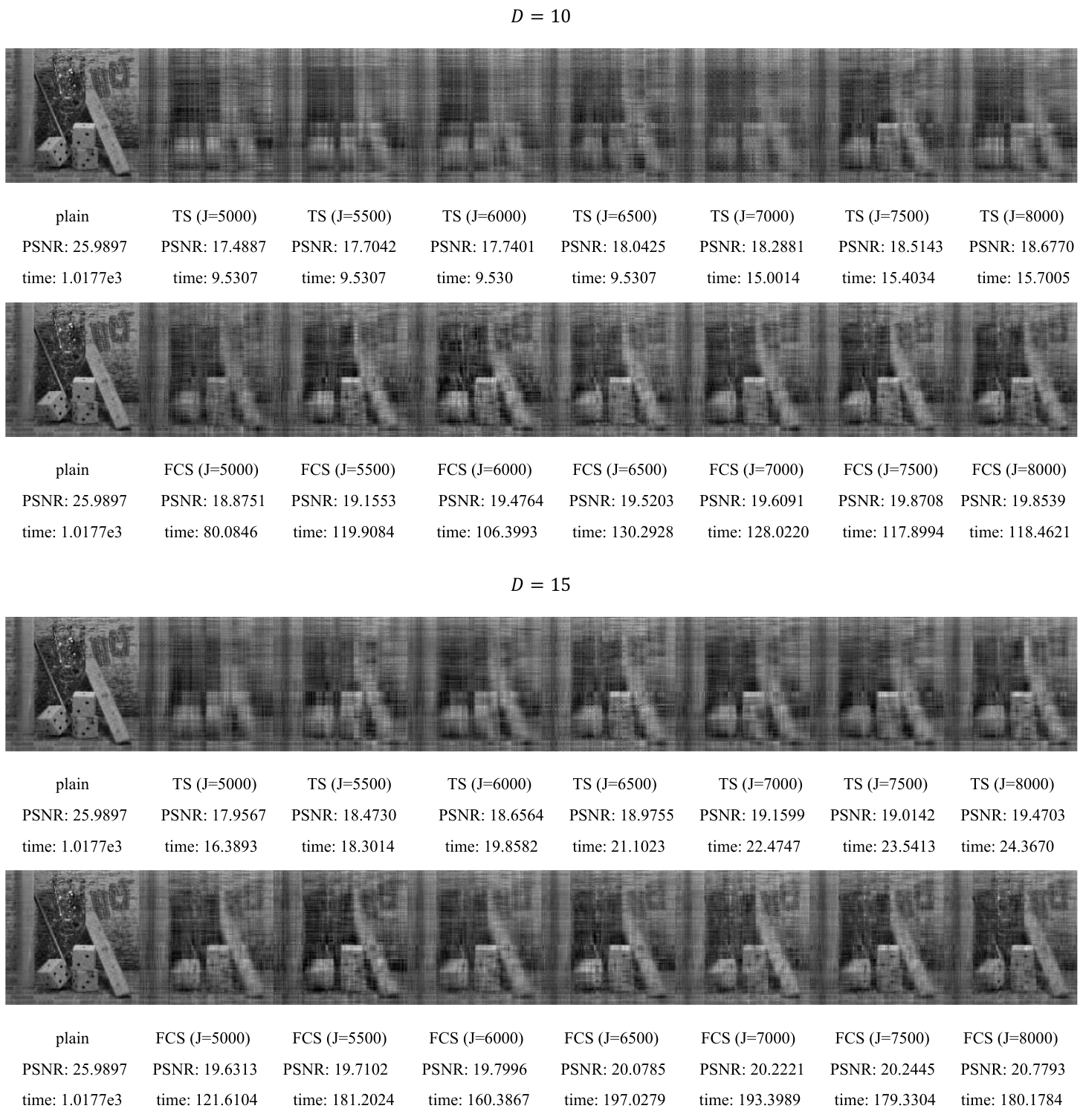}
		\caption{Rank-$30$ approximation of TS-RTPM and FCS-RTPM on light field dataset Buddha. PSNR (dB) and running time (s) are tagged under each approximation figure.}
		\label{fig:Buddha}
	\end{figure}

	We vary the compared Hash lengths from $5000$ to $8000$. The number of independent sketches $D$ is set as $10$ and $15$, respectively. We choose peak signal to noise ratio (PSNR) as the performance metric. We display the $31$st frame of the approximation results of Watercolors in Fig. \ref{fig:Watercolors} and the $1$st frame of the approximation results of Buddha in Fig. \ref{fig:Buddha}, respectively. Clearly FCS-RTPM achieves better approximation quality than TS, especially when the Hash length is small. Although FCS-RTPM is slower than TS-RTPM, it is much faster than the plain RTPM.
	
	\subsubsection{Alternating least squares}
	ALS is another efficient method for CPD, which requires computing \cite{wang2015fast}
		\begin{equation}
			\begin{aligned}
				\mathbf{T}_{(1)}(\mathbf{C}\odot\mathbf{B})&= [\mathcal{T}(\mathbf{I},\mathbf b_1,\mathbf c_1),\cdots,\mathcal{T}(\mathbf{I},\mathbf b_N,\mathbf c_N)]\\
				\mathbf{T}_{(2)}(\mathbf{A}\odot\mathbf{C})&= [\mathcal{T}(\mathbf c_1,\mathbf{I},\mathbf a_1),\cdots,\mathcal{T}(\mathbf c_N,\mathbf{I},\mathbf a_N)]\\
				\mathbf{T}_{(3)}(\mathbf{B}\odot\mathbf{A})&= [\mathcal{T}(\mathbf a_1,\mathbf b_1,\mathbf{I}),\cdots,\mathcal{T}(\mathbf a_N,\mathbf b_N,\mathbf{I})],
			\end{aligned}
		\end{equation}iteratively. Hence we can approximate them like (\ref{eq:T_Ivw}). In the experiment, we generate a synthetic asymmetric CP rank-$10$ tensor $\mathcal{T}=\sum_{r=1}^{10}\mathbf{u}_r\circ\mathbf{v}_r\circ\mathbf{w}_r\in\mathbb{R}^{400\times 400 \times 400}$ as illustrated in the last subsection. We compare the plain, TS and FCS based ALS using the same Hash functions and the results are shown in Table \ref{tab:400synth}. Clearly FCS is more accurate than TS under various conditions. Notice that as the Hash length gets smaller, the accuracy gap between TS and FCS gets bigger, and the speed gap gets smaller.
\begin{table*}[htbp]
	\begin{center}
		\caption{Performance comparison on a synthetic asymmetric CP rank-$10$ tensor $\mathcal{T}\in\mathbb{R}^{400\times400\times400}$ by plain, TS and FCS based ALS.}\label{tab:400synth}
		\scalebox{0.58}{
			\begin{tabular}{ccccccccccccc}
				\toprule
				$\sigma$ &              &             & \multicolumn{5}{c}{Residual norm} & \multicolumn{ 5}{c}{Running time (s)}\\
				\midrule
				\specialrule{0pt}{2pt}{2pt}
				\multirow{11}{*}{0.01} &               & $J$ & 3000          & 4000   & 5000   & 6000   & 7000   & 3000             & 4000     & 5000     & 6000     & 7000     \\
				\specialrule{0pt}{2pt}{2pt}
				& \multirow{4}{*}{\tabincell{c}{TS-\\ALS}}       & $D=10$       & 1.1898 & 0.9063 & 0.7684 & 0.6888  & 0.6198	
				& \textbf{16.0213} & \textbf{15.9789} & \textbf{15.9756} & \textbf{16.6813} & \textbf{16.9187}\\
				\specialrule{0pt}{2pt}{2pt}
				&               & $D=15$    &   0.9721	& 0.7961	& 0.6981	& 0.6272	& 0.5763		& \textbf{16.358}	& \textbf{16.4966}	& \textbf{17.2223}	& \textbf{17.4917}	&\textbf{17.8075}\\
				\specialrule{0pt}{2pt}{2pt}
				&               & $D=20$      & 0.6959	& 0.5899	& 0.5179	& 0.4684	& 0.4337		& \textbf{16.4243}	& \textbf{17.3973}	& \textbf{17.664}	& \textbf{17.9625}	& \textbf{18.3088}\\
				\specialrule{0pt}{2pt}{2pt}
				& \multirow{4}{*}{\tabincell{c}{FCS-\\ALS}}     & $D=10$      & \textbf{0.7801}	&\textbf{0.6429}	&\textbf{0.5618}	&\textbf{0.4915}	&\textbf{0.4547}		&18.3248	&19.6939	&20.9603	&23.3214	&24.7461\\
				\specialrule{0pt}{2pt}{2pt}
				&               & $D=15$      & \textbf{0.6949}	&\textbf{0.5851}	&\textbf{0.5168}	&\textbf{0.4643}	&\textbf{0.4311}		&20.2062	&22.2532	&23.453	&26.2442	&28.9429
				\\
				\specialrule{0pt}{2pt}{2pt}
				&               & $D=20$       &\textbf{0.5122}	&\textbf{0.4342}	&\textbf{0.3877}	&\textbf{0.351}	&\textbf{0.3288}		&22.2656	&24.2863	&26.7042	&30.677	&33.9163
				\\
				\specialrule{0pt}{2pt}{2pt}
				& plain ALS &             & \multicolumn{5}{c}{0.1000}  &\multicolumn{5}{c}{52.2921}\\
				\specialrule{0pt}{2pt}{2pt}
				\midrule
				\specialrule{0pt}{2pt}{2pt}
				\multirow{9}{*}{0.1}
				& \multirow{4}{*}{\tabincell{c}{TS-\\ALS}}       & $D=10$       &1.2927	&1.0001	&0.8641	&0.7749	&0.7119		&\textbf{15.597}	&\textbf{16.2406}	&\textbf{15.908}	&\textbf{16.1767}	&\textbf{17.4988}
				\\
				\specialrule{0pt}{2pt}{2pt}
				&               & $D=15$       &1.0632	&0.8798	&0.7978	&0.7235	&0.6728		&\textbf{16.1131}	&\textbf{16.0404}	&\textbf{16.7699}	&\textbf{17.8976}	&\textbf{17.3802}
				\\
				\specialrule{0pt}{2pt}{2pt}
				&               & $D=20$       &0.7951	&0.6911	&0.6223	&0.5725	&0.5416		&\textbf{16.3257}	&\textbf{17.7667}	&\textbf{18.1851}	&\textbf{18.3393}	&\textbf{18.5008}
				\\
				\specialrule{0pt}{2pt}{2pt}
				& \multirow{4}{*}{\tabincell{c}{FCS-\\ALS}}     & $D=10$       &\textbf{0.8283}	&\textbf{0.7012}	&\textbf{0.6326}	&\textbf{0.5796}	&\textbf{0.5510}		&18.9016	&19.3911	&20.3741	&22.9006	&24.1979
				\\
				\specialrule{0pt}{2pt}{2pt}
				&               & $D=15$       &\textbf{0.7505}	&\textbf{0.6546}	&\textbf{0.5997}	&\textbf{0.5523}	&\textbf{0.5232}		&19.7293	&21.1838	&23.2389	&26.2481	&28.1806
				\\
				\specialrule{0pt}{2pt}{2pt}
				&               & $D=20$       &\textbf{0.5989}	&\textbf{0.5291}	&\textbf{0.4921}	&\textbf{0.4637}	&\textbf{0.4424}		&22.3568	&23.8313	&26.1499	&29.7723	&31.3613
				\\
				\specialrule{0pt}{2pt}{2pt}
				& plain ALS &             & \multicolumn{5}{c}{0.3162} & \multicolumn{5}{c}{53.2564}\\
				\bottomrule
			\end{tabular}
		}
	\end{center}
\end{table*}
	
	\subsection{Tensor regression network compression}
	TRNs replace the last flattening and fully-connected layer of convolutional neural networks by tensor regression layer (TRL) \cite{kossaifi2020tensor}. Here, we show the weight tensor of TRL $\mathcal{W}\in\mathbb{R}^{I_1\times\cdots\times I_N\times C}$ ($C$ represents the number of classes) can be compressed by FCS. Denote $\mathcal{X}\in\mathbb{R}^{B\times I_1\times\cdots\times I_N}$ ($B$ represents batch size) as the input activation tensor. Then the output of TRL is 
		\begin{equation}
			\label{FCS_CP_TRL}
			\mathbf{Y}(i,j)=\left\langle \mathbf{X}_{(1)}(i,:)^{\operatorname{T}}, \mathbf{W}_{(N+1)}(j,:)^{\operatorname{T}} \right \rangle+b.
		\end{equation}for $i\in[B], j\in[C]$. Hence we can approximate (\ref{FCS_CP_TRL}) by
		\begin{equation}
			\hat{\mathbf{Y}}(i,j)=\left\langle {\rm FCS}(\mathbf{X}_{(1)}(i,:)^{\operatorname{T}}), {\rm FCS}(\mathbf{W}_{(N+1)}(j,:)^{\operatorname{T}}) \right \rangle+b,
		\end{equation}which equals the compact form
		\begin{equation}
			\hat{\mathbf{Y}}={{\rm FCS}(\mathbf{X}_{(1)}^{\operatorname{T}})}^{\operatorname{T}}{\rm FCS} (\mathbf{W}_{(N+1)}^{\operatorname{T}}) + \mathbf{b}.
			\label{eq:net_compress}
		\end{equation}The compression ratio (CR) for FCS based TRL is
	$(\prod_{n=1}^{N}I_n) / (\sum_{n=1}^{N}J_n-N+1)=\tilde{I} / \tilde{J}$.
	
	In this experiment, we compare the performance of CS, TS and FCS based TRL on dataset FMNIST \cite{xiao2017fashion} ($C=10$). The TRL model is composed of two convolutional and maxpooling layers. We assume the regression weight tensor admits low-rank CPD (i.e., CP-TRL \cite{cao2017tensor}) with the target CP rank set as $5$. By default, the input activation $\mathcal{X}$ fed to the TRL is of size $B\times7\times7\times32$.
	
\begin{figure}[htbp]
		\centering
		\includegraphics[width=\linewidth]{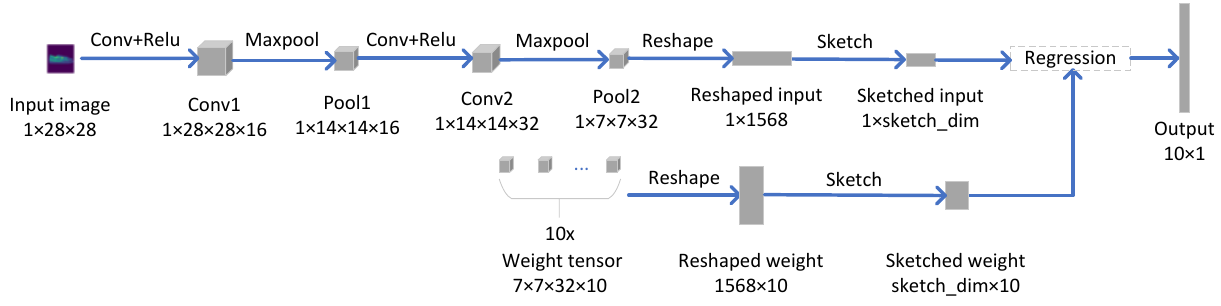}
		\caption{The network structure illustration for the sketching based CP-TRL. The batch size is set to $1$ for convenience of display.}
		\label{fig:network-structure}
	\end{figure}

	We show the network structure in Fig. \ref{fig:network-structure} and comparison results in Table \ref{tab:network-comp}. It can be seen that FCS achieves better classification performance than CS and TS under almost all CRs. 
	\begin{table}[H]
		\centering
		\caption{Classification accuracy by CS, TS and FCS based CP-TRL on dataset FMNIST under various CRs.}
		\scalebox{0.72}{
			\begin{tabular}{ccccccccccc}
				\toprule
				CR &20              &22.22             &25	&28.57	&33.33	&40	&50	&66.67	&100	&200 \\			
				\midrule
				\specialrule{0pt}{2pt}{2pt}
				CS	&\textbf{0.7862}	& 0.7529	& 0.7743	&0.7748	& 0.7451	&0.7333	&0.7555	&0.7322	&0.7732	&0.7104	\\
				\specialrule{0pt}{2pt}{2pt}
				TS	&0.7461	&0.7587	&0.7161	&0.7665	&0.7438	&0.6446	&0.6871	&0.6735	&0.7123	&0.6999		\\
				\specialrule{0pt}{2pt}{2pt}
				FCS	&0.7829	&\textbf{0.8011}	&\textbf{0.7874}	&\textbf{0.7815}	&\textbf{0.7881}	&\textbf{0.7697}	&\textbf{0.7706}	&\textbf{0.7865}	&\textbf{0.7830}	&\textbf{0.7696}		\\
				\bottomrule
			\end{tabular}
		}
		\label{tab:network-comp}
	\end{table}
	
	\subsection{Kronecker product and tensor contraction compression}
	We further compare the performance of FCS against CS and HCS in the Kronecker product compression under the same CRs. We set the independent number of sketches $D=20$ for all sketching methods.
	
	\subsubsection{Kronecker product compression}
\begin{figure}[H]
		\centerline{\includegraphics[width=0.85\columnwidth]{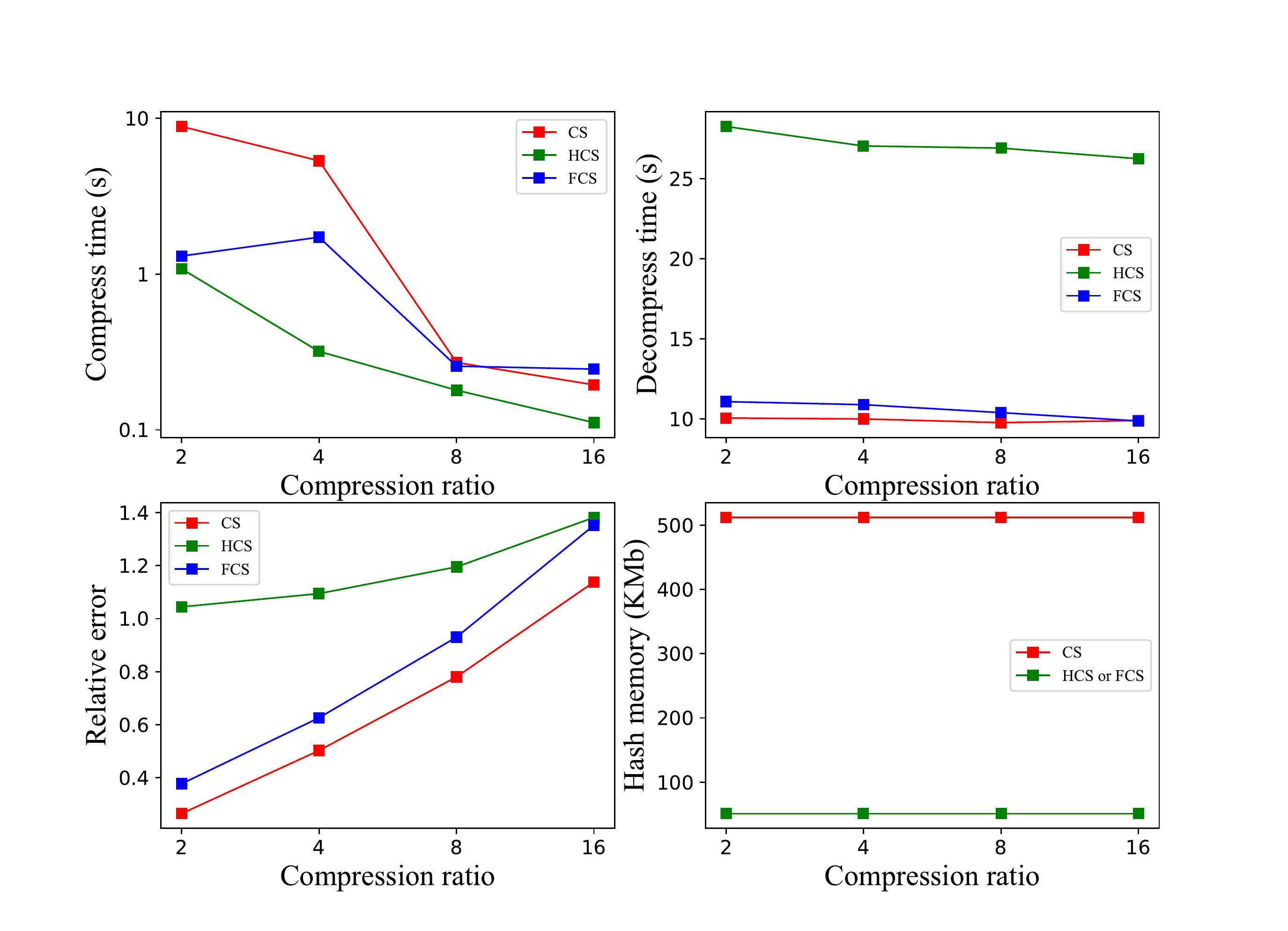}}
		\caption{Compressing time, decompressing time, relative error, and memory for Hash functions for CS, HCS and FCS based Kronecker product.}
		\label{fig:kronecker_product}
	\end{figure}

	For two matrices 
	$\mathbf{A}\in\mathbb{R}^{I_1\times I_2}$, $\mathbf{B}\in\mathbb{R}^{I_3\times I_4}$, we can compress the Kronecker product $\mathbf{A}\otimes\mathbf{B}\in\mathbb{R}^{I_1I_3\times I_2I_4}$ using FCS by:
		\begin{equation}
			\nonumber
			\begin{aligned}
				&{\rm FCS}(\mathbf{A}\otimes\mathbf{B}; \left\{\mathbf{h}_n, \mathbf{s}_n\right\}_{n=1}^{4})\\
				=&\operatorname{F}^{-1}(\operatorname{F}({\rm CS}({\rm vec}(\mathbf{A}); \left\{\mathbf{h}_n, \mathbf{s}_n\right\}_{n=1}^{2}), \tilde{J})*\operatorname{F}({\rm CS}({\rm vec}(\mathbf{B}); \left\{\mathbf{h}_n, \mathbf{s}_n\right\}_{n=3}^{4}), \tilde{J})),
			\end{aligned}
		\end{equation}
	and the decompressing rule is
	\begin{equation}
		\nonumber
		\begin{aligned}
			&\widehat{\mathbf{A}\otimes\mathbf{B}}_{I_3(i_1-1)+i_3, I_4(i_2-1)+i_4}\\
			=&\mathbf{s}_1(i_1)\mathbf{s}_2(i_2)\mathbf{s}_3(i_3)\mathbf{s}_4(i_4){\rm FCS}(\mathbf{A}\otimes\mathbf{B})_{{\rm mod}(\mathbf{h}_1(i_1)+\mathbf{h}_2(i_2)+\mathbf{h}_3(i_3)+\mathbf{h}_4(i_4)-4, 4J-3)+1},
		\end{aligned}
	\end{equation}where $i_n\in[I_n]$, $J$ is the Hash length, $\tilde{J}=4J-3$. We generate two matrices $\mathbf{A}\in\mathbb{R}^{30\times40}$, $\mathbf{B}\in\mathbb{R}^{40\times50}$ with each entry randomly drawn from uniform distribution $[-5, 5]$. We compare the compressing time, decompressing time, relative error, and memory cost for Hash functions of CS, HCS and FCS under various CRs. The comparison results are shown in Fig. \ref{fig:kronecker_product}. Clearly, the compressing time of FCS is shorter than CS when CR is small. Although the compressing time of FCS is longer than CS when CR equals $16$, we argue it is acceptable, since the relative errors of all compared methods are higher than $1$ when CR is $16$, which means the accuracy is even worse than the all-zero recovery. Therefore, in practice we should focus on lower compression ratios with relative errors much smaller than $1$. Besides, although the compressing speed of HCS is faster, it has larger relative error and much longer decompressing time. Finally, the Hash memory required by FCS is only a ten percent of the Hash memory of CS. 
	
	\subsubsection{Tensor contraction compression}
\begin{figure}[H]
	\centerline{\includegraphics[width=0.85\columnwidth]{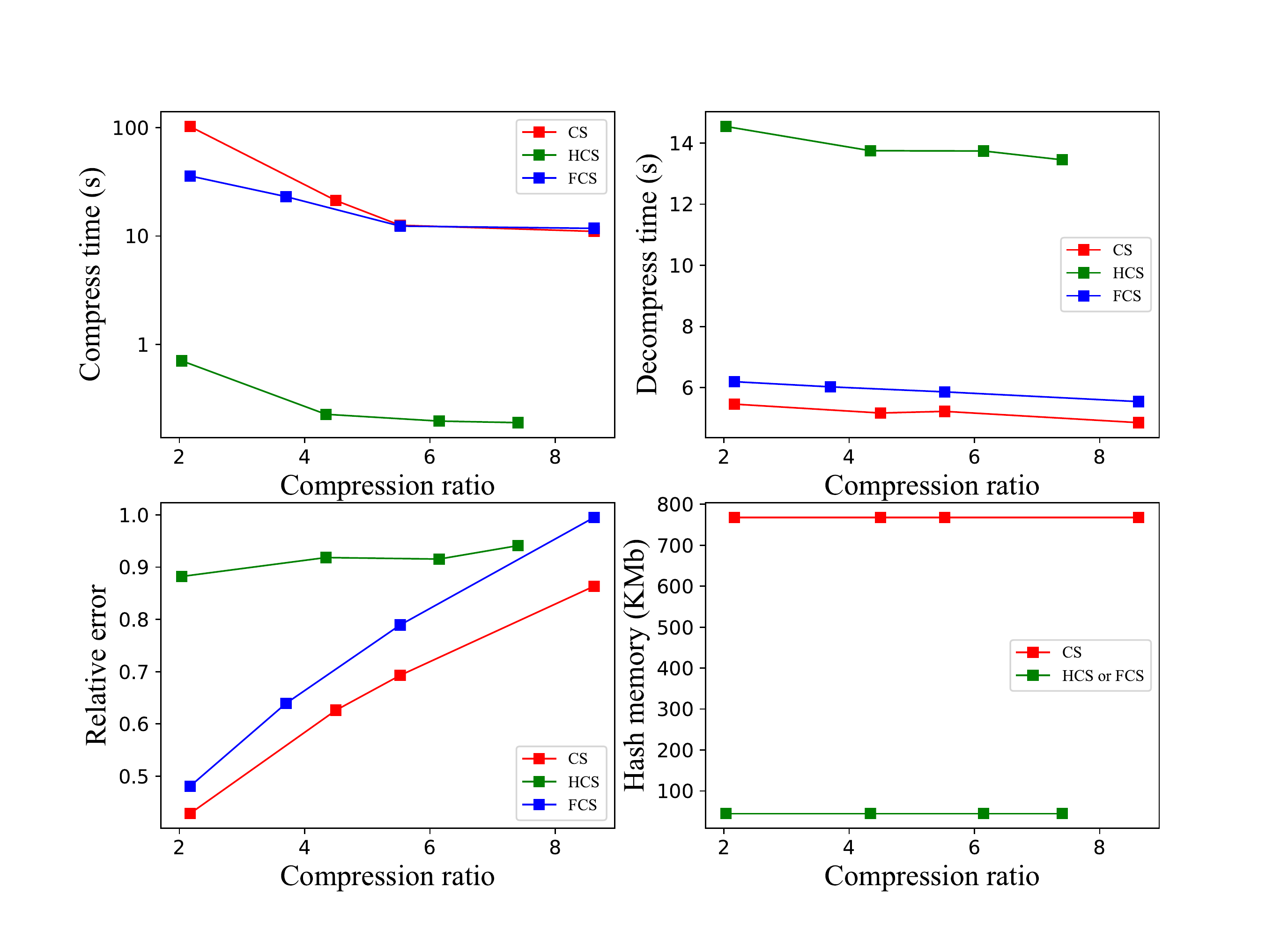}}
	\caption{Compressing time, decompressing time, relative error, and memory for Hash functions for CS, HCS and FCS based tensor contraction.}
	\label{fig:tensor_contraction}
	\end{figure}

	Given two tensors $\mathcal{A}\in\mathbb{R}^{I_1\times I_2\times L}$ and $\mathcal{B}\in\mathbb{R}^{L\times I_3\times I_4}$ with each entry randomly drawn from $[0, 10]$, the tensor contraction on the third mode of $\mathcal{A}$ and first mode of $\mathcal{B}$ produces a tensor $\mathcal{A}\circledcirc_{3, 1}\mathcal{B}\in\mathbb{R}^{I_1\times I_2\times I_3\times I_4}$. By applying FCS, we approximately compress the contraction by:
	\begin{footnotesize}
		\begin{equation}
		\nonumber
		\begin{aligned}
			&{\rm FCS}(\mathcal{A}\circledcirc_{3, 1}\mathcal{B}; \left\{\mathbf{h}_n, \mathbf{s}_n\right\}_{n=1}^{4})\\
			=&\sum_{l=1}^{L}\operatorname{F}^{-1}(\operatorname{F}({\rm CS}({\rm vec}(\mathbf{A}(:,:,l)); \left\{\mathbf{h}_n, \mathbf{s}_n\right\}_{n=1}^{2}), \tilde{J})*\operatorname{F}({\rm CS}({\rm vec}(\mathbf{B}(l,:,:)); \left\{\mathbf{h}_n, \mathbf{s}_n\right\}_{n=3}^{4}), \tilde{J})),\\
		\end{aligned}
	\end{equation}
	\end{footnotesize}
And the decompressing rule is
	\begin{equation}
	\nonumber
	\begin{aligned}
	&\widehat{\mathcal{A}\circledcirc_{3, 1}\mathcal{B}}_{i_1, i_2, i_3, i_4}\\
	=&\mathbf{s}_1(i_1)\mathbf{s}_2(i_2)\mathbf{s}_3(i_3)\mathbf{s}_4(i_4){\rm FCS}(\mathcal{A}\circledcirc_{3, 1}\mathcal{B})_{{\rm mod}(\mathbf{h}_1(i_1)+\mathbf{h}_2(i_2)+\mathbf{h}_3(i_3)+\mathbf{h}_4(i_4)-4, 4J-3)+1},
	\end{aligned}
	\end{equation}where $i_n\in[I_n]$, $J$ is the Hash length, $\tilde{J}=4J-3$. We generate $\mathcal{A}\in\mathbb{R}^{30\times40\times50}$ and $\mathcal{B}\in\mathbb{R}^{50\times40\times30}$ with each entry randomly drawn from $[0, 10]$. The comparison results are shown in Fig. \ref{fig:tensor_contraction}. Again, when CR is small, FCS has faster compressing speed than CS, faster decompressing speed than HCS and higher accuracy than HCS. And the Hash memory is about $5$ percent of CS.
	
	\section{Conclusion}
	A novel sketching method dubbed FCS is presented, which applies multiple shorter Hash functions based CS on the vectorized input tensor. We apply FCS to approximate tensor contraction with the validity guaranteed by theoretical proofs. Numerical experiments on CPD, TRN compression, Kronecker product and tensor contraction compression confirm that FCS achieves competitive approximation quality and running speed compared with various sketching methods.

	\bibliographystyle{elsarticle-num} 
	\bibliography{references_FCS}
	
\end{document}